\pdfoutput=1

\documentclass[11pt]{article}

\usepackage{EMNLP2023}

\usepackage{times}
\usepackage{latexsym}

\usepackage[T1]{fontenc}

\usepackage[utf8]{inputenc}

\usepackage{microtype}

\usepackage{inconsolata}

\usepackage{booktabs}
\usepackage{multirow}
\usepackage{multicol}
\usepackage{graphicx}
\usepackage[linesnumbered,ruled,vlined]{algorithm2e}
\usepackage{subcaption}

\title{Fine-tuning pre-trained extractive QA models for clinical document parsing}

\author{Ashwyn Sharma \\
  Cadence Solutions, USA \\
  \texttt{ashwyn@cadencerpm.com} \\\AND
  David I. Feldman, MD, MPH \\
  Cadence Solutions, USA \\
  Massachusetts General Hospital, Harvard University , USA \\  
  \texttt{david.feldman@cadencerpm.com} \\\AND
  Aneesh Jain \\
  Cadence Solutions, USA \\
  Virginia Polytechnic Institute and State University, USA \\  
  \texttt{aneeshjain70@gmail.com} \\}

\begin{document}
\maketitle
\begin{abstract}
Electronic health records (EHRs) contain a vast amount of high-dimensional multi-modal data that can accurately represent a patient's medical history. Unfortunately, most of this data is either unstructured or semi-structured, rendering it unsuitable for real-time and retrospective analyses. A remote patient monitoring (RPM) program for Heart Failure (HF) patients needs to have access to clinical markers like EF (Ejection Fraction) or LVEF (Left Ventricular Ejection Fraction) in order to ascertain eligibility and appropriateness for the program.
This paper explains a system that can parse echocardiogram reports and verify EF values. This system helps identify eligible HF patients who can be enrolled in such a program. At the heart of this system is a pre-trained extractive QA transformer model that is fine-tuned on custom-labeled data. The methods used to prepare such a model for deployment are illustrated by running experiments on a public clinical dataset like MIMIC-IV-Note. The pipeline can be used to generalize solutions to similar problems in a low-resource setting. We found that the system saved over 1500 hours for our clinicians over 12 months by automating the task at scale.
\end{abstract}

\begin{figure*}
  \centering
  \includegraphics[width=\linewidth,height=0.5\textheight,keepaspectratio]{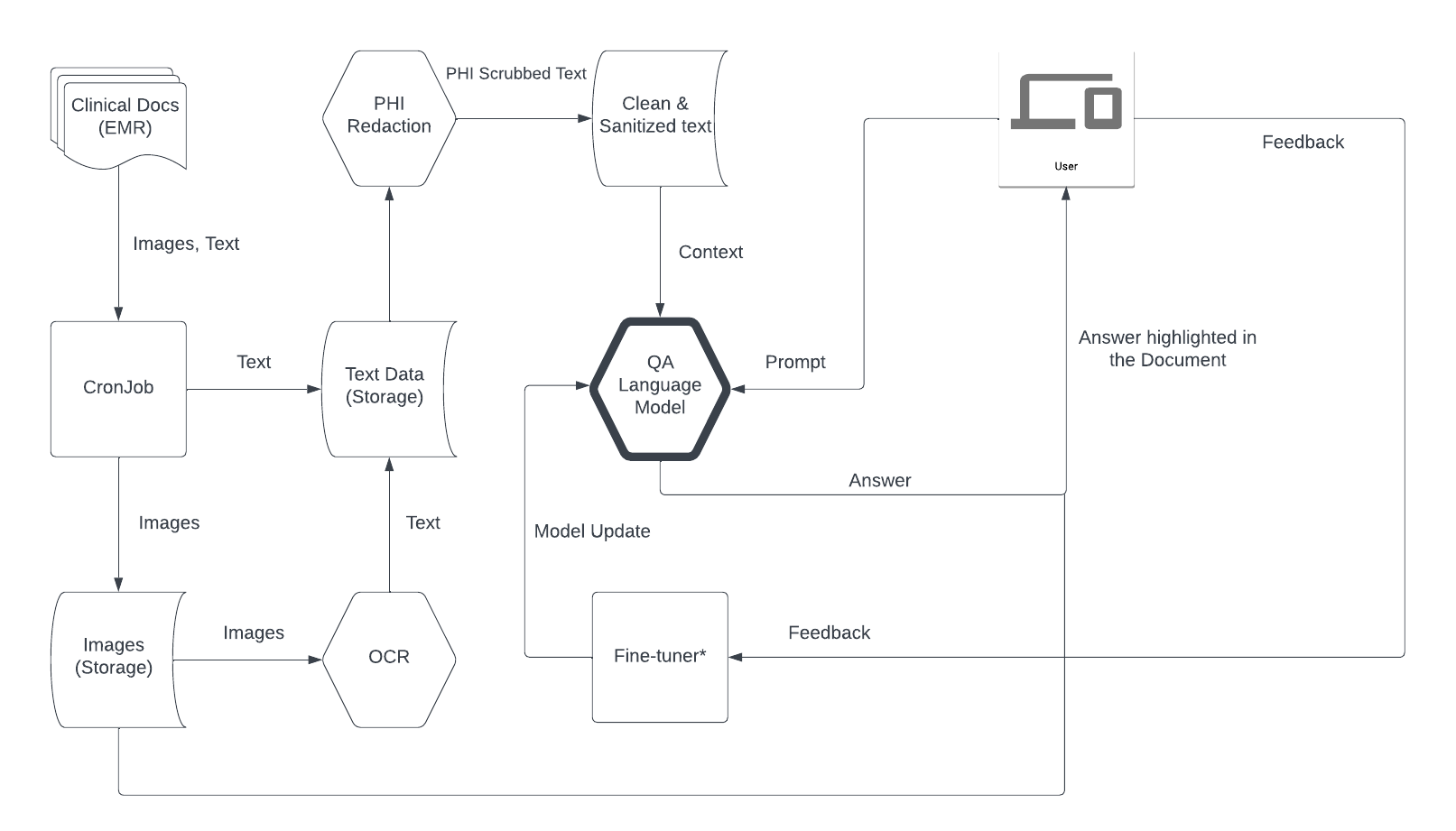}
  \caption{System Architecture for parsing clinical documents using QA models. Fine-tuner is illustrated in Figure \ref{fig:fine-tuner}}
  \label{fig:sys-arch}
\end{figure*}

\section{Introduction}

\begin{figure*}[ht]
\centering
\begin{minipage}[b]{0.45\linewidth}
\centering
\includegraphics[width=\linewidth]{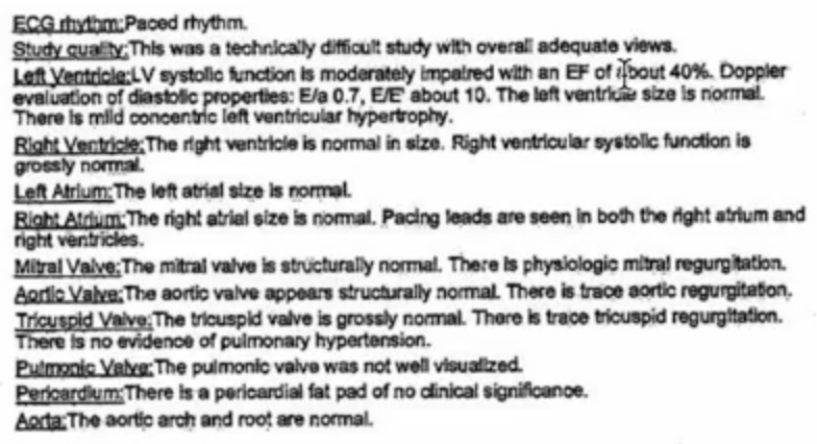}
\caption{Snippet from a sample echocardiogram report (often hard to read for humans).}
\label{fig:echo}
\end{minipage}
\hfill
\begin{minipage}[b]{0.45\linewidth}
\centering
\includegraphics[width=\linewidth]{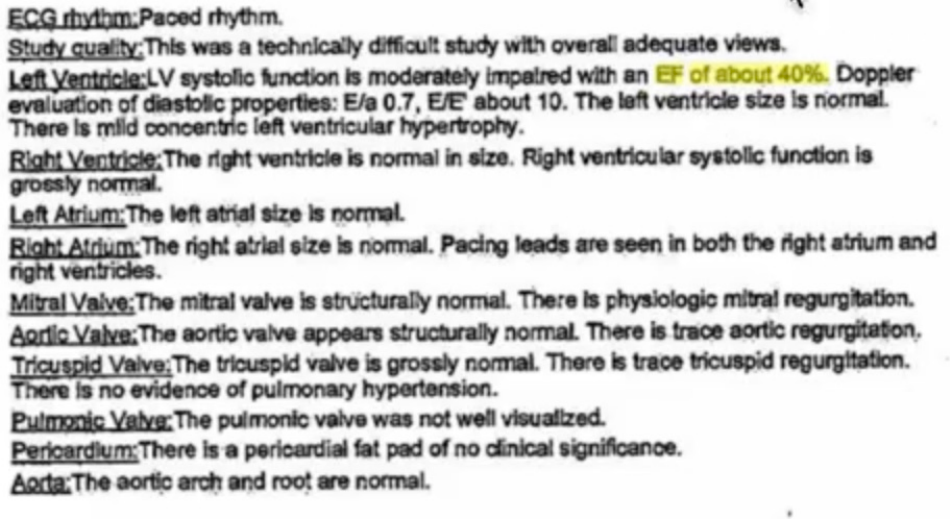}
\caption{Ejection Fraction is highlighted in the PDF report for transparency and feedback.}
\label{fig:echo-highlighted}
\end{minipage}
\end{figure*}

In this paper, we describe a system \cite{CadencePatent} that was designed to assist with identifying HF patients eligible for a Remote Patient Monitoring (RPM) program. Traditional approaches to identifying and enrolling eligible patients into an RPM program depend on patient referrals by clinicians. This method significantly limits how efficiently a provider can convert an eligible, at-risk HF patient into a successful enrollment. Innovative solutions, such as a smart search of provider panels for patients with HF ICD-10 codes, can improve these constraints. Still, many patients go undiagnosed with HF due to the unstructured or semi-structured format of echocardiogram reports (PDFs) and the constellation of HF signs (i.e., labs) and symptoms, which are often hidden in progress notes or discharge summaries (Figure \ref{fig:echo}).

The most critical measure to help determine if HF is present or absent is the reported ejection fraction (EF), which can be found in echocardiogram reports. Ejection fraction is assessed via an echocardiogram using a visual estimate or objective measurement (i.e., single dimension, biplane method of discs). Ejection fraction estimates the percentage of blood pumped to the body with each cardiac cycle. A normal ejection fraction is 50 - 75 percent. Different types of HF are defined by the EF; HF with reduced EF (HFrEF), HF with mildly reduced EF (HFmrEF), and HF with preserved EF (HFpEF) are defined by an EF less than 40\%, an EF of 40-50\% and an EF greater than 50\% with signs and symptoms of HF, respectively\footnote{\url{https://www.pennmedicine.org/updates/blogs/heart-and-vascular-blog/2022/april/ejection-fraction-what-the-numbers-mean}}

The echocardiogram reports are usually PDF-formatted scanned images without a structured section for key information to be tracked (Figure \ref{fig:echo}). Therefore, each echocardiogram requires a clinician to review the text report and formulate a clinical assessment and, when appropriate, a diagnosis of HF \cite{heidenreich20222022, yancy20172017, mcdonagh20212021}. Altogether, this process may take a clinician up to 2 to 3 minutes, significantly limiting the scalability of an RPM program that depends on the rapid diagnosis, enrollment, and treatment of HF patients. The system discussed here was designed to solve this problem and ensure the program has the data it needs to accurately identify appropriate patients for enrollment. We discuss the implementation details in Section \ref{sec:sys-arch} of this paper.

The solution to this problem is centered on a standard natural language processing (NLP) task - \textit{extractive question answering}, the task of identifying spans of text in a blob of text (also known as context) that answers a question about the context. The advent of transformers \cite{vaswani2017attention} and the accessibility of pre-trained large language models (LLMs) via repositories like HuggingFace \cite{wolf2020transformers} have significantly facilitated the rapid development of sophisticated text-based and natural language-oriented systems. Many such models have achieved state-of-the-art performance on extractive QA tasks \cite{lewis2019mlqa}. Although these models perform well on open-domain inputs, they need to be adapted to the clinical domain \cite{yoon2020pre} by leveraging the transfer learning technique of fine-tuning. In this case, we fine-tuned the \textit{distilbert-base-cased-distilled-squad} model checkpoint\footnote{\url{https://huggingface.co/distilbert-base-cased-distilled-squad}} from HuggingFace \cite{wolf2020transformers} on a custom-labeled dataset of echocardiogram reports achieving near-perfect performance in extracting EF values from them. Since the real dataset can not be shared publicly, we illustrate the methods and concepts used by running experiments on discharge summaries sampled from the MIMIC-IV-Note \cite{johnsonmimic, goldberger2000physiobank} dataset. We hope the same pipeline can be used by teams constrained for resources to bootstrap solutions to similar problems.

By the time of writing this paper, the pipeline has been deployed for 12 months and has potentially saved an estimated total of over 1500 clinician hours. Reducing time spent on such non-clinical tasks can free clinicians to focus on patient care, which includes initiating and titrating life-saving medical therapy for HF patients \cite{heidenreich20222022, yancy20172017, mcdonagh20212021}. The platform allows clinicians to report incorrectly parsed EF values when identified; however, we have not received notification of such reports since the pipeline's deployment. The authors recognize that echocardiogram reports are formatted very consistently for a given laboratory, and the fact that the model saw similar distributions during training and testing may have contributed to the high accuracy of the pipeline. However, accuracy evaluation with datasets exhibiting a distribution shift also yields high numbers with these methods, albeit not as perfect as our deployment - as we can see with our MIMIC-IV-Note \cite{johnsonmimic, goldberger2000physiobank} experiments.

\begin{figure}
  \centering
  \includegraphics[width=\linewidth,height=0.5\textheight,keepaspectratio]{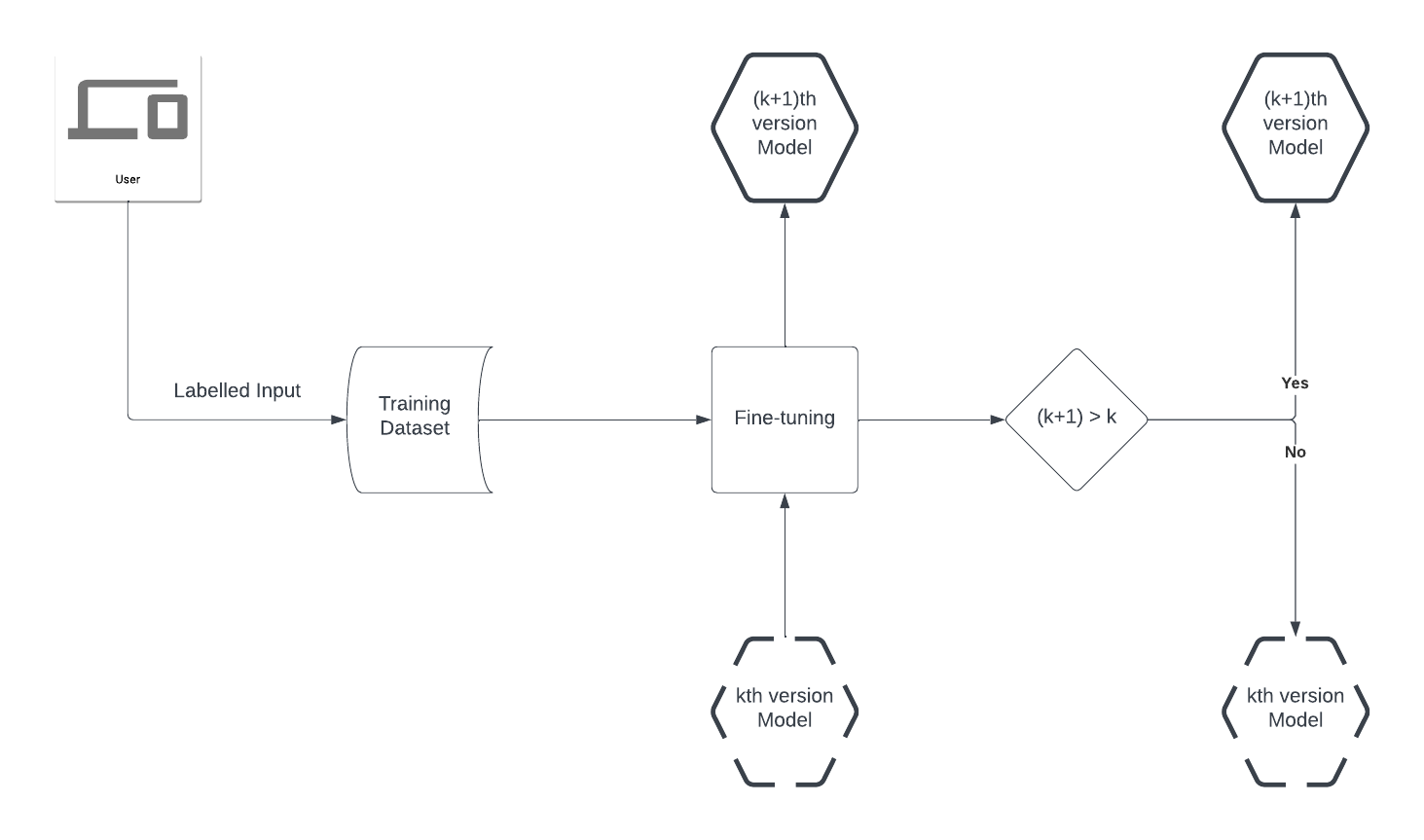}
  \caption{Fine-tuner module of the deployed pipeline updates the model if fine-tuning on feedback improves the metrics.}
  \label{fig:fine-tuner}
\end{figure}

\begin{algorithm*}
    \label{algo1}
    \SetKwInOut{Input}{Input}
    \SetKwInOut{Output}{Output}
    \Input{Discharge summary notes from MIMIC-IV-Note}
    \Output{Results on the impact of fine-tuning a QA model for EF extraction.}
    Set random seed for reproducibility\;
    Set the list of EF keywords = [EF, LVEF, Ejection Fraction]\;
    Find discharge summary notes containing at least one of the EF keywords\;
    Randomly sample 100 notes\;
    Split into training and testing datasets with an 80/20 ratio.\;
    Label EF values in each of the 100 notes\;
    Set the list of prompts that expect the same answer (Ejection Fraction)\;
    Use a pretrained QA model to answer each question in the prompt list and evaluate the EM accuracy and F1 score on the test notes\;
    Fine-tune the QA model using the training split for each question in the prompt list\;
    Compare the test set performance metrics (EM accuracy and F1 score) of the fine-tuned model with that of the pretrained model\;
    Repeat the above steps with other random seeds\;
    \caption{EF Identification using a QA Model - the \textit{label-and-fine-tune} pipeline.}
\end{algorithm*}

\section{Related Work}

The modeling methods we used leverage recent advancements in extractive QA models powered by better architectures \cite{vaswani2017attention}, availability of pre-trained LLMs in public repositories \cite{wolf2020transformers}, and open-domain datasets like SQuAD curated for extractive QA \cite{rajpurkar2018know}.
We recognize that prior work on clinical domain adaptation of QA models \cite{xie2022extracting, soni2020evaluation} helped us make informed choices when designing the system and setting up the experiments. We make a note of such choices and cite the papers in the following sections when appropriate.

\citet{xie2022extracting}'s work on extracting seizure frequency and date of the last seizure from epilepsy clinic notes is highly relevant and solves a similar problem from an NLP perspective. However, there are significant differences worth noting. They focused on a retrospective study of non-public data and evaluated their models against human annotators. In contrast, this paper focuses on a system deployed in the real world and illustrates the methods by reproducible experimentation on a public dataset.

\section{System Architecture} \label{sec:sys-arch}

The system architecture for the deployed pipeline is illustrated in Figure \ref{fig:sys-arch}. The guiding principles for this design were security, scalability, and maximum transparency. As seen in Figure \ref{fig:sys-arch}, echocardiogram reports are fetched from the EHR and then converted to text using an Optical Character Recognition (OCR) API like AWS Textract \footnote{\url{https://aws.amazon.com/textract/}}. Text processed by OCR is then fed to a PHI (Protected Health Information) redaction API like AWS Comprehend Medical \footnote{\url{https://docs.aws.amazon.com/comprehend-medical/latest/dev/textanalysis-phi.html}} to limit the volume of data with PHI in the pipeline. Ejection fractions are then extracted from the PHI-redacted text by the model described in Section \ref{sec:models}.

Since QA models output the text span containing the required EF value, index information from the span output is combined with pixel-level bounding box output from the OCR API to highlight the pixels in the original PDF (Figure \ref{fig:echo}) where the answer was found (Figure \ref{fig:echo-highlighted}). The highlighted PDF is made available to the clinicians for review and feedback. This bi-directional feedback loop is critical to the transparency aspect of the system.

\section{Experiments and Methods}

\subsection{Datasets}

The model used in the pipeline described in Section \ref{sec:sys-arch} is fine-tuned and tested on a dataset of echocardiogram reports with EF values as labels. A sample echocardiogram report is shown in Figure \ref{fig:echo}. Since actual patient data can not be made publicly available, we ran experiments on the MIMIC-IV-Note \cite{johnsonmimic, goldberger2000physiobank} dataset using the same methods. The dataset contains 331,794 de-identified discharge summaries, of which 75,107 contain keywords like EF, LVEF, and Ejection Fraction.

\subsection{Models} \label{sec:models}

We chose to fine-tune the \textit{distilbert-base-cased-distilled-squad} model \footnote{\url{https://huggingface.co/distilbert-base-cased-distilled-squad}} - specifically, the pre-trained checkpoint available on HuggingFace \cite{wolf2020transformers}. This model is a fine-tuned checkpoint of \textit{DistilBERT-base-cased} \cite{sanh2019distilbert}, fine-tuned using (a second step of) knowledge distillation on SQuAD v1.1 \footnote{\url{https://huggingface.co/datasets/squad}}. Given its low memory footprint yet high performance reported on the GLUE language understanding benchmark, the checkpoint was considered the most reasonable choice by the authors. We chose to skip the evaluation of pre-trained clinical QA models because Soni et al. \cite{soni2020evaluation} showed in their evaluation that models fine-tuned on an open-domain dataset like SQuAD perform better on new clinical QA tasks than models fine-tuned on medical-domain datasets like CliCR \cite{vsuster2018clicr} and emrQA \cite{pampari2018emrqa}.

\subsection{Experiments} \label{sec:exp}

Algorithm 1 lists the steps involved in reproducing the experiments on the MIMIC-IV-Note dataset \cite{johnsonmimic, goldberger2000physiobank}. First, for each iteration of the experiment, we set a random seed for determinism, which is used by Python packages like \textit{random}, \textit{numpy}, and \textit{torch}. We then filter the discharge summary notes from the MIMIC-IV-Note dataset \cite{johnsonmimic, goldberger2000physiobank} based on a set of EF keywords. Then, 100 notes are sampled, split into a train/test split of 80/20, and labeled by human annotators to create a golden dataset. The pre-trained QA model (\textit{distilbert-base-cased-distilled-squad} \footnote{\url{https://huggingface.co/distilbert-base-cased-distilled-squad}}) is then fine-tuned on the golden dataset against the top three prompts recommended by cardiologists on our team. Note that all three prompts expect the same answer (EF). Finally, the performance of the fine-tuned model is compared against the pre-trained model on the test set. We choose a small test set size of 20 to simulate the lack of available annotated data in a real industrial setting and maximize the amount of training data seen by the model. These steps were repeated with four random seeds, and results were reported in Section \ref{sec:results}. 

The experiments were designed to be reproducible and reusable as a general \textit{label-and-fine-tune} pipeline on similar QA tasks. Our code allows researchers to sample from MIMIC-IV-Note \cite{johnsonmimic, goldberger2000physiobank}, annotate the sampled dataset using label studio \cite{LabelStudio}, and fine-tune the pre-trained model using the Trainer API from HuggingFace \cite{wolf2020transformers}. 

\begin{table*}
\centering
\begin{tabular}{@{}lccc@{}}
\toprule
\textbf{Prompt} & \textbf{Model} & \textbf{F1} & \textbf{EM} \\ 
\midrule
What is the EF percentage? & pre-trained & 57.05 & 52.94 \\
& finetuned & \textbf{92.64} & \textbf{86.76} \\
\midrule
What is the ejection fraction? & pre-trained & 17.35 & 13.23 \\
& finetuned & \textbf{96.56} & \textbf{92.64} \\
\midrule
What is the systolic function? & pre-trained & 1.96 & 0 \\
& finetuned & \textbf{93.13} & \textbf{88.23} \\
\bottomrule
\end{tabular}
\caption{Results from fine-tuning the QA model on MIMIC-IV-Note samples. Metrics are averaged over four runs with different random seeds.}
\label{tab:results}
\end{table*}

\section{Results} \label{sec:results}

Table \ref{tab:results} presents the results of the experiments described in Section \ref{sec:exp}. We can see that fine-tuning the pre-trained QA model on a custom-labeled dataset improves average EM (Exact Match) accuracy by \textbf{33\%,  79\%, }and\textbf{ 88\%} (absolute percentage points) for the three prompts. Similarly, F1 score improves by \textbf{35\%,  79\%, }and\textbf{ 91\%} (absolute percentage points) . The metrics used were as defined in the original SQuAD paper \cite{rajpurkar2016squad}. 

Note how the pre-trained model struggles with the third prompt, \textit{"What is the systolic function?"}, an indirect query that does not include direct references to any EF keywords. We can see that fine-tuning helps the model handle different prompts robustly - the standard deviation for both metrics  improves by \textbf{90.69\%} on average with fine-tuning (Table \ref{tab:sensitivity}). This reduced sensitivity to prompts can prevent teams from spending crucial time and effort on prompt engineering.

\begin{table}[h]
\centering
\begin{tabular}{@{}lcc@{}}
\toprule
\textbf{Metric} & \textbf{EM} & \textbf{F1} \\
\midrule
Pre-trained & 27.55 & 28.42\\
Fine-tuned & \textbf{3.06} & \textbf{2.13}\\
\bottomrule
\end{tabular}
\caption{Prompt sensitivity: Standard deviation of metrics across the three prompts that expect the same answer (Ejection Fraction).}
\label{tab:sensitivity}
\end{table}

\section{Limitations}

The authors recognize that the described system and experiments have limitations despite the successful deployment and promising results. In theory, it is tough for our pipeline to recover from errors made by the OCR API. An end-to-end visual QA model\cite{antol2015vqa} or a multi-modal generative model like GPT-4\cite{openai2023gpt4} might address that problem, and we leave the exploration of such models for future work. We also chose to skip domain adaption of the base model with masked language modeling (MLM) because  \citet{xie2022extracting} reported that MLM domain adaption was not always helpful. Also, our experiments leveraged dataset annotations by non-clinicians only. We were encouraged by the findings from  \citet{xie2022extracting}, where they found no difference in the correctness of the non-clinician and clinician annotators.

\section{Conclusion}

In this work, we described a real-world application of extractive QA models in the context of an RPM program. We laid out the system architecture of the deployed pipeline and reported results from the illustrative experiments on the MIMIC-IV-Note \cite{johnsonmimic, goldberger2000physiobank} dataset. Results clearly demonstrated the value of fine-tuning pre-trained models on a dataset that closely represents the domain of the clinical application. In addition to improving accuracy, fine-tuning also reduced sensitivity to different prompts, which is critical in a clinical setting. 
A major obstacle to using clinical NLP models in the real world is the presence of PHI in the data, which makes external APIs like GPT-4\cite{openai2023gpt4} impractical. This warrants the development of in-house solutions, which is a non-viable option for teams with limited resources. We hope that our \textit{label-and-fine-tune} pipeline can help such teams design sophisticated solutions to their NLP problems involving clinical text. 

\bibliography{anthology,custom}
\bibliographystyle{acl_natbib}

\end{document}